\documentclass[letterpaper, 10 pt, conference]{ieeeconf}  

\IEEEoverridecommandlockouts                              

\usepackage{graphics} 
\usepackage{epsfig} 
\usepackage{times} 
\usepackage{amsmath} 
\usepackage{amssymb}  
\usepackage{float}
\usepackage{booktabs}
\usepackage{caption}
\usepackage{lipsum}
\usepackage{graphicx}
\usepackage{multirow}
\usepackage{xcolor}  
\usepackage{subfigure}
\usepackage{xspace}
\usepackage{threeparttable}
\usepackage{color}
\usepackage{enumerate}
\usepackage{subcaption}
\usepackage{bbding}
\usepackage{capt-of}
\usepackage{cite}
\usepackage{soul}
\title{\LARGE \bf
VTAO-BiManip: Masked Visual-Tactile-Action Pre-training \\ with Object Understanding for Bimanual Dexterous Manipulation
}

\author{Zhengnan Sun$^{1}$, Zhaotai Shi$^{1}$, Jiayin Chen$^{1}$, Qingtao Liu$^{1}$, Yu Cui$^{1}$, Jiming Chen$^{1}$ and Qi Ye$^{1\dag}$ 
\thanks{$^{1}$College of Control Science and Engineering, Zhejiang University, Hangzhou, 310027, China}
\thanks{$^{\dag}$Qi Ye (Corresponding author, qi.ye@zju.edu.cn) is with the College of Control Science and      Engineering and the State Key Laboratory of Industrial Control Technology, Zhejiang University, and also with the Key Key Lab of CS\&AUS of Zhejiang Province.}%
}

\begin{document}
{
\newcommand{\reflabel}{dummy} 


\newcommand{\seclabel}[1]{\label{sec:\reflabel-#1}}
\newcommand{\secref}[2][\reflabel]{Section~\ref{sec:#1-#2}}
\newcommand{\Secref}[2][\reflabel]{Section~\ref{sec:#1-#2}}
\newcommand{\secrefs}[3][\reflabel]{Sections~\ref{sec:#1-#2} and~\ref{sec:#1-#3}}

\newcommand{\eqlabel}[1]{\label{eq:\reflabel-#1}}
\renewcommand{\eqref}[2][\reflabel]{(\ref{eq:#1-#2})}
\newcommand{\Eqref}[2][\reflabel]{(\ref{eq:#1-#2})}
\newcommand{\eqrefs}[3][\reflabel]{(\ref{eq:#1-#2}) and~(\ref{eq:#1-#3})}

\newcommand{\figlabel}[2][\reflabel]{\label{fig:#1-#2}}
\newcommand{\figref}[2][\reflabel]{Fig.~\ref{fig:#1-#2}}
\newcommand{\Figref}[2][\reflabel]{Fig.~\ref{fig:#1-#2}}
\newcommand{\MyFigref}[2][\reflabel]{Fig.~\ref{figs/#1-#2}}
\newcommand{\figsref}[3][\reflabel]{Figs.~\ref{fig:#1-#2} and~\ref{fig:#1-#3}}
\newcommand{\Figsref}[3][\reflabel]{Figs.~\ref{fig:#1-#2} and~\ref{fig:#1-#3}}

\newcommand{\tablelabel}[2][\reflabel]{\label{table:#1-#2}}
\newcommand{\tableref}[2][\reflabel]{Table~\ref{table:#1-#2}}
\newcommand{\Tableref}[2][\reflabel]{Table~\ref{table:#1-#2}}
\newcommand{\etal}{et al.}
\newcommand{\eg}{e.g.}
\newcommand{\ie}{i.e. }
\newcommand{\etc}{etc. }

\def\bfmu{\mbox{\boldmath$\mu$}}
\def\bftau{\mbox{\boldmath$\tau$}}
\def\bftheta{\mbox{\boldmath$\theta$}}
\def\bfdelta{\mbox{\boldmath$\delta$}}
\def\bfphi{\mbox{\boldmath$\phi$}}
\def\bfpsi{\mbox{\boldmath$\psi$}}
\def\bfeta{\mbox{\boldmath$\eta$}}
\def\bfnabla{\mbox{\boldmath$\nabla$}}
\def\bfGamma{\mbox{\boldmath$\Gamma$}}

%
%


\newcommand{\R}{\mathbb{R}}

\newcommand{\be}{\begin{equation}}
\newcommand{\ee}{\end{equation}}

\newcommand{\todo}[1]{\textcolor{red}{TODO: #1}}
\newcommand{\yq}[1]{\textcolor{black}{#1}}
\newcommand{\szn}[1]{\textcolor{black}{#1}}
\newcommand{\lqt}[1]{\textcolor{cyan}{\textbf{lqt:}\xspace#1}}
\newcommand{\method}{VTAO-BiManip\xspace}
\newcommand{\muldata}{VTAO\xspace}
\newcommand{\del}[1]{\textcolor{gray}{Delete:#1}}
\newcommand{\md}[1]{\textcolor{orange}{\textbf{modified: }\xspace#1}\xspace}

\newcommand\blfootnote[1]{
\begingroup 
\renewcommand\thefootnote{}\footnote{#1}
\addtocounter{footnote}{-1}
\endgroup 
}

\twocolumn[{%
\renewcommand\twocolumn[1][]{#1}%
\maketitle
\begin{center}
    \captionsetup{type=figure}
    \includegraphics[width=\textwidth]{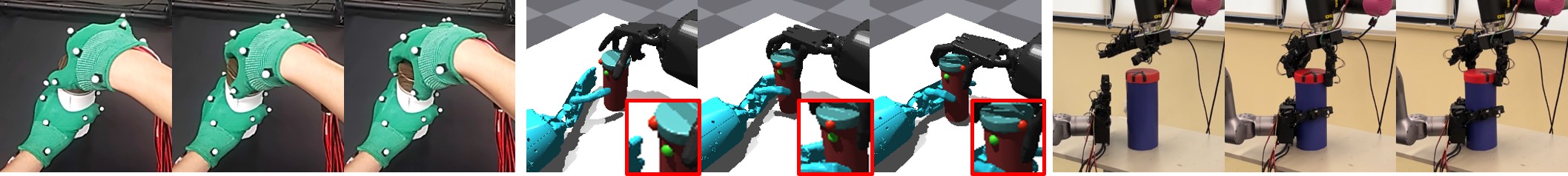}
    \captionof{figure}{Left: examples of our bottle-cap turning dataset. Middle: simulation results of the bottle-cap task with Shadow Hand~\cite{shadowhand}. Right: the real-world deployment of the bottle-cap task with Leap Hand~\cite{shaw2023leaphand}.}
    \label{fig:title-teaser}
\end{center}%
}]

\thispagestyle{empty}
\pagestyle{empty}
\blfootnote{$^{1}$College of Control Science and Engineering, Zhejiang University, Hangzhou, 310027, China}
\blfootnote{$^{\dag}$Qi Ye (Corresponding author, qi.ye@zju.edu.cn) is with the College of Control Science and Engineering and the State Key Laboratory of Industrial Control Technology, Zhejiang University, and also with the Key Key Lab of CS\&AUS of Zhejiang Province.}
\blfootnote{This work was supported in part by NSFC under Grants (No. 62088101, 62233013, 62293511) and  Key Research and Development Program of Zhejiang Province (No. 2025C01072).}

\begin{abstract}

Bimanual dexterous manipulation remains a significant challenge in robotics due to the high DoFs of each hand and their coordination. 
Existing single-hand manipulation techniques often leverage human demonstrations to guide RL methods but fail to generalize to complex bimanual tasks involving multiple sub-skills.
In this paper, we propose \method, a novel framework that integrates visual-tactile-action pre-training with object understanding, aiming to enable human-like bimanual manipulation via curriculum reinforcement learning (RL).
We improve prior learning by incorporating hand motion data, providing more effective guidance for dual-hand coordination. 
Our pretraining model predicts future actions as well as object pose and size using masked multimodal inputs, facilitating cross-modal regularization.
To address the multi-skill learning challenge, we introduce a two-stage curriculum RL approach to stabilize training.
We evaluate our method on a bimanual bottle-cap twisting task, demonstrating its effectiveness in both simulated and real-world environments. 
Our approach achieves a success rate that surpasses existing visual-tactile pretraining methods by over 20\%.

\end{abstract}



\section{Introduction}

Humans effortlessly perform diverse complex bimanual tasks like opening containers and assembling furniture in daily life via dynamic inter-hand coordination. Such capabilities rely on three fundamental \szn{components}: 
1) kinematic inter-hand coordination: dynamically adjusting joint-level movements between hands to maintain operational stability, 
2) real-time object-state sensing: modifying manipulation strategies based on real-time state changes of manipulated objects, 
3) temporally composition of complementary skill primitives: executing phase-dependent primitives (grasping, holding, and rotating) under physical constraints. 
While humans perform these tasks effortlessly, robotic dexterous bimanual manipulation remains an open challenge. 
The control complexity grows exponentially when coordinating two high-DoF hands (totaling 30+ joints) under physical constraints, particularly in dynamic scenarios \szn{with coupled dynamics between object states and manipulation requirements.}

Recent advances in dexterous manipulation~\cite{Rajeswaran2018DAPG, qin2022dexmv, wu2022ILAD, liu2024m2vtp} have demonstrated progress in single-hand scenarios by leveraging human demonstrations and multi-modal sensing. Particularly, Liu et al.~\cite{liu2024m2vtp} established a visual-tactile pre-training framework that enables transferable manipulation skills through masked autoencoding. 
However, these methods fundamentally rely on constrained task setups - for instance, fixing objects on tables to bypass bimanual coordination challenges. Directly extending such frameworks to dynamic bimanual tasks like bottle-cap twisting reveals three irreducible challenges:
1) kinematic coordination \szn{limitation}: while vision and binary tactile sensing aids single-hand finger coordination~\cite{liu2024m2vtp}, they fail to capture inter-hand kinematic synergy critical for dynamic bimanual manipulation.
2) object-state decoupling: when the bottle tilts during cap rotation, the passive hand must respond based on real-time 3D pose changes. Current methods lack \szn{explicit perception of object pose and size}, leading to incoherent hand movements under dynamic states.
3) sub-skill composability: successful cap twisting requires temporally aligned sub-skills - the passive hand must establish stable grasping before the active hand initiates rotation. However, single-stage policies in prior work~\cite{liu2024m2vtp} lack mechanisms to coordinate these interdependent sub-processes. 

To address the issues, in this paper, we propose \method, a novel framework extending~\cite{liu2024m2vtp} through \szn{two key} innovations: visual-tactile-action pre-training method with object understanding \yq{for multi-modal perception} and curriculum reinforcement learning \yq{for action control policy learning}. 
To address kinematic coordination blindness and  object-state decoupling, we introduce hand action and object-centric understanding as complementary modalities during pre-training: 
1) while prior work relies on visual-tactile fusion, we introduce hand joint angles as explicit kinematic observables and predict future actions to model inter-hand synergy. This enables the policy to dynamically coordinate bimanual motions; for instance, progressively adjusting the \szn{finger joint angles of the passive hand} to counteract active-hand torque during cap twisting.
Notably, unlike imitation learning which predicts next-step actions from current states, our action pre-training predicts future action sequences using masked multi-modal inputs. 
2) we also augment the pre-training framework with object pose and size estimation, forcing the encoder to learn geometry-aware representations. By reconstructing the bottle's 6D pose and 3D dimensions from visual-tactile-action inputs, the model gains explicit awareness of dynamic state changes. During deployment, this allows real-time adaptation.

To address sub-skill composability, a two-stage curriculum reinforcement learning framework is proposed. In the first stage, the bottles are fixed on the table, encouraging the left hand to grab the bottle and the right hand to unscrew the cap. This phase reduces exploration complexity, building foundational sub-skills of both hands. In the second stage, the bottle is physically released, forcing the two-hand cooperative manipulation skill learning. 

In summary, our contributions are as follows:

\begin{itemize}
    \item We propose \method, a novel visual-tactile-action pre-training method with object understanding for bimanual dexterous manipulation. This approach leverages human data to pre-train models for action prediction and object state estimation. 
    \item We propose a curriculum reinforcement learning framework to address the challenge of multi-skill learning within a single task.
    \item \szn{We propose a bimanual bottle-cap twisting task where our VTAO-BiManip pre-training and curriculum RL framework succeeds both in simulation and the real world.} 
\end{itemize} 

\section{Related Works}


\subsection{Multimodal Pretraining for Robotics}
Pre-training methods that use unsupervised learning~\cite{radford2021clip, he2022mae, oord2018representation} have emerged as particularly effective in multimodal representation generalization.
Due to the availability of extensive manipulation datasets that include visual and language modalities~\cite{goyal2017sth_sth_v2, grauman2022ego4d, Damen2022epic_kitchen}, some researches~\cite{ma2023liv, nair2022r3m, karamcheti2023voltron, xiao2022mvp} explore the pretraining representation of these modalities in robotic tasks, demonstrating their effectiveness and generalization capabilities.
~\cite{brohan2023rt2} extends this by integrating action modalities with vision and language, enabling robots to predict and execute tasks.~\cite{wen2023atm} improves downstream performance by predicting future motions of any point via visual pertaining. 
However, pre-training on action modalities has primarily focused on simple end-effectors, with limited exploration in dexterous manipulation.
For tactile, studies such as~\cite{chen2023VTT, mejia2024hearing, gano2024vt, liu2024m2vtp} highlight the effectiveness of visuo-tactile fusion in robotic manipulation. However, its application in bimanual tasks remains underexplored.
To address this gap, we propose a novel approach that incorporates multimodal pre-training, including action prediction, into bimanual manipulation.


\subsection{Robotic Bimanual Manipulation}

Learning-based bimanual manipulation has gained significant attention in robotics. 
Although some studies have successfully used demonstration data~\cite{stepputtis2022system, zhao2023aloha, kim2021transformer}, the lack of high-quality multi-fingered hand demonstrations often restricts their applicability to simple end-effectors. 
Efforts to collect such data for multi-fingered hands have been made~\cite{handa2020dexpilot, qin2022one, arunachalam2023dexterous}, but challenges such as retargeting errors and motion retargeting latency limit their practical use. 

Sim-to-real approaches have demonstrated success in robotic manipulation~\cite{akkaya2019solving, liu2023dexrepnet, qin2023dexpoint, qi2023hand}, including deformable object manipulation~\cite{matas2018sim}, open doors~\cite{rajeswaran2017DAPG}, and in-hand rotations~\cite{chen2023visual_dexterity}. 
However, most methods focus on single dexterous hands~\cite{chen2023sequential, yin2023RotateWithoutSeeing, bao2023dexart, pitz2023dextrous} or dual-gripper systems~\cite{li2023efficient, lin2023bi_touch}.  
Some progress has been made in dual-dexterous-hand manipulation~\cite{chen2023bi, huang2023dynamic_handover, zakka2023robopianist, Lin2024twisting}.~\cite{chen2023bi} construct a dexterous bimanual manipulation platform, offering 20 manipulation tasks.~\cite{huang2023dynamic_handover} employ multi-agent RL to train a bimanual handover task, incorporating a trajectory prediction model to bridge the sim-to-real gap.~\cite{zakka2023robopianist} presents a system in which robot hands are trained to play the piano using DRL.\szn{~\cite{Lin2024twisting} explores RL-based twisting policies for predefined cylindrical containers using sparse visual keypoints, demonstrating limited zero-shot transfer through contact-aware rewards.}
However, these works mainly rely on visual and proprioceptive modalities, which somewhat limit the performance of manipulation tasks.
Multimodal data has proven effective in dual-gripper~\cite{hogan2020tactile, mejia2024hearing, sunil2023visuotactile_cloth} and single dexterous hand tasks~\cite{yuan2024RobotSynesthesia, qi2023inhand-visuotactile, guzey2024see}, its application in bimanual dexterous manipulation remains largely unexplored. 
Our work represents a pioneering effort to apply multimodal sensing to dual-dexterous-hand manipulation.


\section{Method}
In this section, we first introduce our data collection system to capture ego-centric human manipulation data, which includes \underline{V}ision, \underline{T}actile, \underline{A}ction, and \underline{O}bject (\muldata) information (see \secref[Method]{System}). \secref[Method]{Pretrain} then introduces \method, a novel pre-training framework that utilizes human manipulation priors for multimodal feature extraction and fusion. Finally, \secref[Method]{RL} details the implementation of a dexterous bimanual bottle cap unscrewing task in simulation, followed by policy training through the curriculum RL.

\subsection{\muldata Capture System for Human Bi-Manipulation} \label{sec:Method-System}
We develop a \muldata data collection system to capture multimodal data during human bimanual manipulation. \szn{The visual and tactile acquisition methods are directly adopted from our previous system~\cite{liu2024m2vtp}. The key addition is a motion capture system that records hand motion trajectories and bottle cap poses in the same coordinate system.} Below, we detail this \muldata capture system: 

\subsubsection{\textbf{System Overview}}
The \muldata capture system includes: 1) Hololens2~\footnote{https://www.microsoft.com/en-us/hololens/buy} for visual data capture, 2) tactile and motion acquisition gloves for both hands, 3) the CHINGMU motion capture system~\footnote{https://www.chingmu.com/} with 16 cameras to capture precise hand movements and 6-DoF object pose, and 4) a personal computer responsible for data acquisition and alignment, as shown in \figref[method]{CapSystem}.
The CHINGMU system plays a critical role in acquiring high-quality bimanual motion trajectories and object pose data, which are essential for the following model pre-train.
Each acquisition glove consists of two layers: an inner layer with 20 tactile sensors (identical to the design in~\cite{liu2024m2vtp}), and an additional outer layer with retroreflective markers for optical motion capture.

\begin{figure}[t!]
    \centering
    \includegraphics[width=0.9\linewidth]{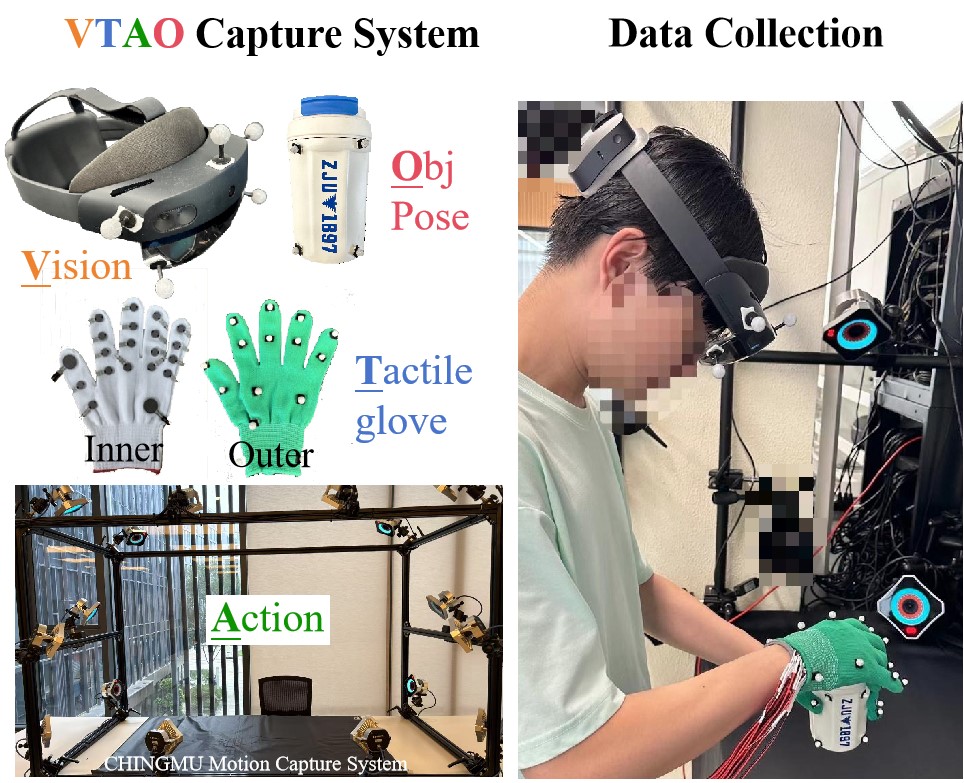}
    \caption{Our \muldata capture system for dual-hand human manipulation.}
    \label{fig:method-CapSystem}
    \vskip -0.7cm
\end{figure}

\subsubsection{\textbf{Data Alignment and Unified Representation}}

Multimodal data are captured independently, with visual data sampled at 30 Hz, tactile data at 200 Hz, and mocap data at 1 kHz. 
Data from other modalities are synchronized to the visual modality through local timestamps recorded by the personal computer during acquisition.
\szn{To unify the motion capture and visual coordinate systems, we install markers on both the Hololens2 and a calibration board, then simultaneously capture their positions in the motion capture system while acquiring the calibration board's location in the visual system, thereby calibrating the transformation between the camera coordinates and the Hololens2's rigid-body coordinates.}
This transformation enables mapping of the 6-DoF bottle pose and hand motion trajectories to the RGB camera coordinate system of the visual modality, to facilitate model comprehension.

\subsubsection{\textbf{Retarget Human Manipulation Trajectories}} \label{sec:Method:System-Retarget}
To unify bimanual action representations across pre-training and downstream tasks, we retarget human bimanual trajectories to dexterous robotic hands via nonlinear optimization, following DexRepNet~\cite{liu2023dexrepnet}. The objective, aligned with~\cite{handa2020dexpilot}, preserves geometric similarity by minimizing distances between wrist-to-fingertips and wrist-to-phalanxes across all motion phases, ensuring natural robotic hand poses, and is defined as:
\begin{equation}
   q_t^{\mathbf{R} *} = \underset{q_t^{\mathbf{R}}}{\operatorname{argmin}} 
                                                \sum_{i=0}^{N}  \| 
                                                       \mathbf{v}^{\mathbf{R}}_i  \left (  M^{\mathbf{R}}, q_t^{\mathbf{R}} \right )  
                                                       -  \mathbf{v}_i^{\mathbf{H}} \left ( M^{\mathbf{H}}, q_t^{\mathbf{H}}  \right )
                                                 \|^{2} ,
\end{equation}
where $\mathbf{v}_i^{\mathbf{R}}$ and  $\mathbf{v}_i^{\mathbf{H}}$ denote the target vectors for the robotic and human hands respectively, defined as wrist-to-fingertip and wrist-to-phalanx distance vectors. These vectors are computed via forward kinematics of the dual Shadow Hand $M^\mathbf{R}$ and human hand model $M^\mathbf{H}$, using joint angles $q^\mathbf{R}_t$ and $q^\mathbf{H}_t$ at timestep $t$. 
The optimized joint angles $q_t^{\mathbf{R} *}$ are obtained through this minimization process.
We optimize each timestep of the collected human bimanual trajectories to generate retargeted robotic motion sequences, which are subsequently used for robotic policy pre-training.

\begin{figure*}[t]
    \centering
    \includegraphics[width=0.95\textwidth]{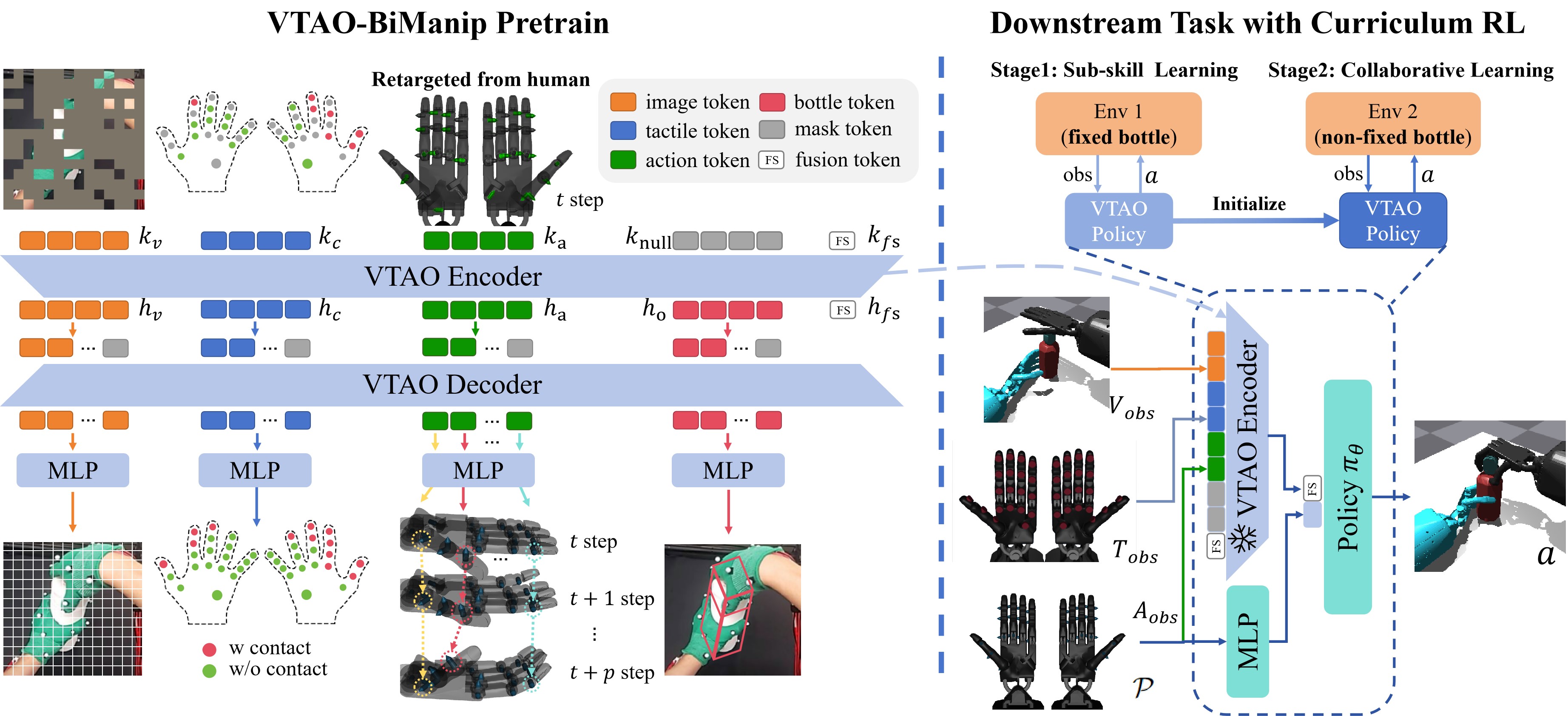}
    \caption{Overview of our \method pre-train framework and its downstream bimanual manipulation with sub-skill and collaborative curriculum reinforcement learning.}
    \label{fig:method-Pipeline}
    \vskip -0.2cm
\end{figure*}

\subsubsection{\textbf{\szn{VTAO Dataset}}}

Using our \muldata acquisition system, we collect 216 human bimanual manipulation trajectories with 26 different bottles, spanning 6-17 seconds per trajectory and totaling 61,684 temporally aligned multimodal frames.
Through preprocessing, RGB images are cropped to $224\times 224$ resolution, tactile signals are binarized with a $\lambda=0.4V$ threshold, bottle poses are aligned to the camera coordinate system, and human hand motions are kinematically retargeted to dual robotic hands.
Each processed frame includes a RGB image ($V \in \R^{224\times 224\times 3}$), bimanual tactile signals ($T \in \R^{40} $, 20 tactile sensors per hand), robotic joint angles of the current step ($A \in \R^{48}$, 24 joints per hand), and an object state vector ($O \in \R^{9}$) encoding the 6-DoF pose (position and orientation) and 3-D size information (width, height, depth) of the bottle.

\subsection{Masked \muldata Transformer for Pre-Training} \label{sec:Method-Pretrain}
To enhance dual-handed robotic skill learning through multimodal data fusion, we propose \method, a MAE-based framework pre-trained on our collected \muldata dataset.
As shown in \figref[method]{Pipeline}, \yq{our pre-training framework follows an encoder-decoder structure: 1) a multimodal encoder processes visual, tactile, and current action inputs for each time step from the demonstration trajectories, projecting them into latent tokens that integrate information for different modalities via attention; 2) a decoder reconstructs the original sensory modalities, predicts subsequent actions, and estimates object states from these latent tokens.}
This joint reconstruction mechanism inherently enables cross-modal feature fusion during pre-training, where the model learns to recover complete environmental interactions from partial sensory observations.
\szn{Notably, the object state information is only used for VTAO-BiManip pre-training and is not incorporated into RL policy learning during downstream tasks to maintain practical feasibility.}


\subsubsection{\textbf{\muldata Encoder} $E_\theta (V, T, A) \rightarrow (h_{\text{fs}}, h_v, h_c, h_a,\allowbreak h_o)$} 
\szn{The VTAO encoder processes a single frame input (V, T, A), comprising an RGB image $V$, binarized tactile signals $T$, and action $A$. The action $A$ is represented by robotic joint angles retargeted from human hand motions.} 
Notice that the object information $O$ is excluded from inputs due to deployment constraints, but will be reconstructed via mask tokens appended to the input sequence.
\yq{For all the input and output modalities, they are tokenized and each token is a vector of $d$ dimensions.}
The visual input ($V \in \R^{224 \times 224 \times 3}$) is divided into 196 non-overlapping patches via a Conv2D network to extract patch tokens $k_v\in \R^{196 \times d}$.
\szn{Binarized tactile signals $T \in \R^{40}$ (20 tactile sensors per hand) are projected into latent tokens $k_c \in \R^{40 \times d}$ via an MLP, establishing a one-to-one mapping between tactile sensors and tokens.}
This binarization follows the design in~\cite{liu2024m2vtp}, aiming to reduce the sim-to-real gap and eliminate the need for precise calibration across taxels.
\szn{The action $A \in \R^{48}$ (joint angles from 24 joints per hand) is encoded into action tokens $k_a \in \R^{48 \times d}$ through a separate MLP, with each token representing one joint angle.}
All input tokens are concatenated into a sequence $(k_{\text{fs}}, k_v, k_c, k_a, k_{\text{null}})$, where $k_{\text{fs}} \in \mathbb{R}^{d}$ is the fusion token (a special token aggregating global information for downstream tasks) and $k_{\text{null}} \in \mathbb{R}^{9 \times d}$ are mask tokens for $O$ reconstruction. 
Following positional encoding and MAE-style masking~\cite{he2022mae}, we randomly mask portions of tokens from different modalities: visual tokens $k_v$ (ratio $r_v \in [0, 1)$), tactile tokens $k_c$ ($r_c \in [0, 1)$), action joint tokens $k_a$ ($r_a \in [0, 1)$) and object mask tokens $k_\text{null}$ ($r_o \in [0, 1)$). 
\yq{The multi-modal tokens,  with fusion token,} are encoded by a Transformer~\cite{kim2021transformer}, outputting latent representations $(h_{\text{fs}}, h_v, h_c, h_a, h_o)$ for downstream tasks.

\subsubsection{\textbf{\muldata Decoder} $D_\theta (h_v, h_c, h_a, h_o, m) \rightarrow (\widehat{V}, \widehat{T},\allowbreak [\widehat{A}, \widehat{A_{+1}}, ..., \widehat{A_{+p}}], \widehat{O})$}
To enhance the feature extraction ability of the \muldata encoder $E_\theta$ for downstream tasks, we define four reconstruction objectives for the decoder: 
\textit{a)} reconstruct the image $\widehat{V}$; \textit{b)} reconstruct tactile signals $\widehat{T}$; \textit{c)} reconstruct current actions $\widehat{A}$ and predict future actions $[\widehat{A_{+1}}, ..., \widehat{A_{+p}}]$ for $p$ steps; 
\textit{d)} reconstruct the 6-DoF pose and 3-D size information of the bottle $\widehat{O}$, as shown in \figref[method]{Pipeline} (left). 
Here, $m$ denotes the mask tokens used to fill in missing dimensions of the input sequence.

\paragraph{\textbf{Reconstruct Image $\widehat{V}$ and Tactile Signals $\widehat{T}$}}
We employ a Transformer decoder and MLP layers to reconstruct the image $\widehat{V}$ and tactile signals $\widehat{T}$ from the corresponding tokens $h_v$ and $h_c$, respectively. \szn{This approach has been demonstrated to benefit downstream applications in~\cite{liu2024m2vtp}.}

\paragraph{\textbf{Reconstruct Current Actions and Predict Future Actions $[\widehat{A}, \widehat{A_{+1}}, ..., \widehat{A_{+p}}]$}}
To enable the encoder $E_\theta$ to acquire prior knowledge of human manipulation tasks from single-frame \muldata input, we use a Transformer decoder and an MLP layer to reconstruct the current dual-hand action $\widehat{A}$ and predict future actions $[\widehat{A_{+1}}, ..., \widehat{A_{+p}}]$ for p steps, based on the action tokens $h_a$.

\paragraph{\textbf{Reconstruct Bottle Informations $\widehat{O}$}} \label{sec:Method:Pretrain-Decoder:ReBottle}
Our framework reconstructs the 6-DoF pose of the assembled object (bottle with an attached cap) relative to the camera's 3D coordinate frame, along with its 3D bounding box dimensions (width, height, depth). This is achieved by decoding the object tokens $h_\text{o}$, which forces the encoder $E_\theta$ to learn geometry-dynamic representations. These representations encode structural priors of the object and its spatiotemporal state transitions, thereby enabling real-time motion adaptation during dynamic manipulation.

\paragraph{\textbf{Loss Design}}
We formulate the loss function as:
\begin{align}
         L(\theta) = &W_{\text{img}}\cdot L (V, \widehat{V}) + 
                               W_{\text{tac}} \cdot L(T, \widehat{T}) +  
                               W_{\text{bot}} \cdot L(O, \widehat{O}) \nonumber \\
                               &W_{\text{act}} \cdot L([A, A_{+1}, ...,  A_{+p}], [\widehat{A}, \widehat{A_{+1}}, ..., \widehat{A_{+p}}]) ,
\end{align}
where $W_{\text{img}}, W_{\text{tac}}, W_{\text{bot}}$, and $W_{\text{act}}$ are weights for each loss term, set to 1, 2, 5, and 2, respectively. 
$L(\cdot)$ denotes the Euclidean distance between two vectors.

\subsection{\szn{Curriculum RL for Dexterous Bimanual Manipulation}} \label{sec:Method-RL}

\subsubsection{\textbf{Task}}
To assess the effectiveness of \muldata pre-training on downstream tasks, we implement a bimanual bottle-cap rotation task (focusing solely on the unscrewing process). The setup positions the bottle body at the center of the table with the left hand in a pre-grasp position (left side) and the right hand positioned above it, as shown in the submitted video. The bottle body and cap remain connected through a z-axis rotational joint. Successful operation is defined by cap rotation exceeding 4 rad (angular displacement) rather than physical removal, \szn{as defined in~\cite{liu2024m2vtp}}.

\subsubsection{\textbf{Problem Formulation}}
We model the bimanual manipulation task as a Markov Decision Process (MDP) defined by a tuple $(\mathcal{S}, \mathcal{A}, \mathcal{T}, \mathcal{R}, \gamma)$. Here, $\mathcal{S}$ and $\mathcal{A}$ represent the state space and action space. The policy $\pi_\theta: \mathcal{S} \rightarrow \mathcal{A}$ maps states to actions. Transition dynamics are given by $\mathcal{T}:\mathcal{S} \times \mathcal{A} \rightarrow \mathcal{S}$, and the reward function by $\mathcal{R}:\mathcal{S} \times \mathcal{A} \rightarrow \mathbb{R}$. The discount factor $\gamma\in\left ( 0, 1 \right ] $. Our objective is to maximize the expected discounted reward $J(\pi)=\mathbb{E}_{\pi}\left[\sum_{t=0} \gamma^{t} r\left(s_{t}, a_{t}\right)\right]$ to train a policy network $\pi_{\theta}$. We employ the PPO algorithm to facilitate the learning of manipulation skills. Our RL architecture is illustrated in \figref[method]{Pipeline} (right).

\subsubsection{\textbf{State Space with Multi-Modal Pre-trained Representations} $\mathcal{S} $}
We define the state space as $\mathcal{S} = \{V_{\text{obs}}, T_{\text{obs}}, \mathcal{P}\}$, where $V_{\text{obs}}$ is the ego-centric RGB image, $T_{\text{obs}}$ is the binary tactile signal from both hands, thresholded by $\lambda$, and $\mathcal{P}$ is proprioception data of both hands, including joint angles and velocities. 
\szn{Notably, object state information is excluded from the input state space $\mathcal{S}$ during RL training, as such privileged data is difficult to directly measure under real-world deployment constraints.} 
To maintain feature representation from pre-training and improve stability in RL with sparse rewards, we freeze the parameters of the pre-trained \muldata model during RL training. 
The final feature vector input to the policy network is $\{h_\text{fs}, \phi(\mathcal{P})\}$, where $h_\text{fs} = \mathrm{E}_{\theta_{f}}(V_\text{obs}, T_\text{obs}, A_\text{obs})$, with $\mathrm{E}_{\theta_{f}}$ representing the pre-trained \muldata encoder with frozen parameters.
$A_\text{obs} \in \mathcal{P}$ represents the joint angles of both hands. 
$h_\text{fs}$ is the FS token output by $\mathrm{E}_{\theta_{f}}$. 
$\phi(\cdot)$ is a linear layer.

\subsubsection{\textbf{Action Space} $\mathcal{A}$}
In the simulation, we use Shadow Hand~\cite{shadowhand} as the manipulator. This five-fingered robotic hand features 24 DoFs. Excluding the thumb, the distal joints of the four fingers are tendon-driven, resulting in a total of 20 DoFs. We implement dual-hand manipulation, where the right hand is controlled with its initial position serving as the reference point, while the left hand is freely manipulated with six DoFs. Consequently, the action $a=\pi_{\theta}(s) \in \mathbb{R}^{46}$.

\subsubsection{\textbf{Learning Curriculum with Stage-specific Reward Design $\mathcal{R}$}}
It is challenging to train precise dual-hand manipulation with non-fixed objects from scratch. 
\szn{In our experiments, single-stage training yields a consistent zero success rate due to insufficient exploration in the high-dimensional bimanual action space.}
To overcome this and stabilize the learning process, we implement a progressive two-stage curriculum with phase-dependent reward functions $r_n (n\in \{1, 2\})$, where each stage's total reward combines the individual rewards of the left and right hands: $r_n=r_{\text{left}_n}+r_{\text{right}_n}$.

\textbf{Stage 1: Sub-skill Learning.}
In the first stage, we fix the bottle body on the table, enabling the left hand to learn to grasp the bottle and the right hand to learn to unscrew the cap. 
The reward function of the first stage $r_1$ is:
\begin{itemize}
    \item Left hand learns bottle grasping through:
    \begin{equation}
        r_{\text{left}_1} = \underbrace{\alpha_1 d_{\text{h2b}}}_{\text{distance penalty}} + \underbrace{\alpha_2 N_{\text{con}}}_{\text{contact reward}}.
    \end{equation}
    \noindent \szn{Here, $\alpha_1\!=\!-5$, $\alpha_2\!=\!0.05$. $d_{\text{h2b}}$ is the distance between the left hand and the bottle. $N_{\text{con}}$ represents the number of contacts between the left hand and the bottle body.}
    
    \item Right hand develops cap rotation skills via:
    \begin{equation}
        r_{\text{right}_1} = \underbrace{\alpha_3\min(a_c,4.0)}_{\text{rotation angle reward}} + \underbrace{\alpha_4 C_{\text{flag}}v_c}_{\text{rotation speed reward}} + \underbrace{\alpha_5 e^{-10d_{\text{fz}}}}_{\text{approach reward}}. 
    \end{equation}
    \noindent Here, $\alpha_3\!=\!0.5$, $\alpha_4\!=\!1.1$, $\alpha_5\!=\!0.5$. $a_{\text{c}}$ is the rotation angle of the bottle cap. $C_{\text{flag}}$ equals 1 if the right hand is in contact with the cap, and 0 otherwise. $v_{\text{c}}$ is the rotation speed of the cap. The approach reward term encourages the right-hand fingertips to approach the bottle cap, with $d_{\text{fz}}$ denoting the z-axis distance between the fingertips and cap's top surface.
\end{itemize}

\textbf{Stage 2: Collaborative Learning.}
In the second stage, we release the bottle body, allowing both hands to collaborate in unscrewing the cap.
The rewards for the second stage, $r_2$ are as follows:
\begin{itemize}
    \item Left hand manages grasping and stabilization:
    \begin{equation}
        r_{\text{left}_2} = \underbrace{\beta_1 e^{-5d_{\text{h2t}}}}_{\text{hand approach}} + \underbrace{\beta_2 e^{-10d_{\text{o2i}}}}_{\text{bottle stability}} + \underbrace{\frac{\beta_3}{|d_{\text{qua}}|+1}}_{\text{upright maintenance}}.
    \end{equation}
    \noindent Here, $\beta_{1}\!=\!\beta_2\!=\!\beta_3\!=\!1.0$. $d_{\text{h2t}}$ is the distance between the left hand and the target position $p_t = (x_0, y_0 - 0.06, z_0)$, with $(x_0,y_0,z_0)$ being the hand's initial position. The bottle stability term encourages the bottle to remain near its initial position, with $d_{\text{o2i}}$ denoting the distance between the bottle's current and initial position. The upright maintenance term encourages the axis of the bottle remains perpendicular to the table, where $d_{\text{qua}} = 2 \cdot \arcsin\left(\min\left(\lVert q_{\text{diff}} \rVert_2, 1.0\right)\right)$, and $q_{\text{diff}}= q_{\text{bot}} \cdot q_{\text{ini}}$ represents the vector part of the quaternion difference between the bottle's current quaternion $q_{\text{bot}}$ and its initial orientation $q_{\text{ini}}$. 
     
    \item Right hand maintains stage 1 rewards $r_{\text{right}_2} = r_{\text{right}_1}$ while adapting to dynamic bottle movements.
\end{itemize}

\section{Experiment Results}

In this section, we design experiments to validate: 1) the effectiveness of multimodal fusion pre-training; 2) the efficacy of action prediction; and 3) the validity of object understanding. 

\subsection{Experiment Settings}

\subsubsection{\textbf{Metrics}}
We use \textbf{success rate} to evaluate the effectiveness of our method on downstream tasks. 
A task is considered successful only if the dexterous hand rotates the bottle cap by more than 4 rad.
\szn{During training, we compute the success rate for each iteration as the moving average over the preceding ten episodes across all parallel environments.}
For evaluation, we select the model from the 3500th iteration of the second training stage and measure the average success rate across 10 manipulations for each bottle in the evaluation set.


\subsubsection{\textbf{Implementation Details}}
\szn{We deploy our bimanual bottle cap twisting task in the Isaac Gym simulation environment~\cite{makoviychuk2021isaac}.}
We select 15 bottles from ShapeNet~\cite{chang2015shapenet} with varying body and cap sizes, using 10 for training (Seen) and 5 for testing (Unseen).  
Our \method runs on a system with an Intel Xeon Gold 6326 and an NVIDIA 3090. 
During pretraining, we use the AdamW optimizer with a learning rate of 2e-5. 
The mask ratios are set to $r_v = 0.75$, $r_c = 0.5$, and $r_a = 0.5$.
Training with a minibatch size of 8 takes about 2 hours for 300 epochs.
For RL, the two-stage curriculum learning takes approximately 62 hours to learn the dual-hand manipulation strategy. In the first stage, we train the policy using PPO for 1000 iterations. The resulting model is then used as the initialization for the second stage, where we further train for 3500 iterations.

\subsubsection{\textbf{Baselines}}
To validate the effectiveness of our method, we design multiple baseline models with different configurations, as summarized in Table 1. The baselines follow a naming convention of \textbf{VTAO}, where each capital letter indicates the inclusion of a specific pretrained modality using the MAE architecture shown in Figure 3: V-Visual (RGB observations), T-Bimanual binary tactile signals, A-Bimanual action joints (current reconstruction + 5-frame prediction) and O - Bottle information reconstruction. The absence of a letter denotes exclusion of the corresponding modality.
\yq{We also construct a baseline that trains the multi-modal representation and the action policy from scratch by end-to-end reinforcement learning \textbf{VTA-Scr}. The baseline employs a ResNet encoder for visual inputs and MLP networks for processing tactile and action modalities.}

\begin{table}[t!]
  \centering
      \vspace{0.1cm}
      \caption{Settings for baselines. }  
      \tabcolsep=2.0pt
      \begin{tabular}{l | cccc}
      \toprule
      \textbf{Methods} & \textbf{Modality} & \textbf{Pretrain} & \textbf{JointPretrain} & \textbf{ReconBottle}    \\
      \midrule
       \textit{VTA-Scr} 
         & \textit{V+T+A} & $\times $        &  $ \times  $        &   $\times $     \\
       \textit{V}           
         & \textit{V} & $ \checkmark $    &  $\times$        & $\times$             \\
        \textit{T} 
         & \textit{T} &  $ \checkmark $        &  $\times$        & $\times$ \\
        \textit{A} 
         & \textit{A} &  $ \checkmark $       &  $\times$       & $\times$ \\
        \textit{VT} 
         & \textit{V+T} &  $ \checkmark $      &  $ \checkmark $       &  $\times$           \\
        \textit{VTA} 
         & \textit{V+T+A} & $ \checkmark $    & $ \checkmark $        &  $\times$        \\
        \textit{VTO}
         & \textit{V+T}   & $ \checkmark $    & $ \checkmark $        & $ \checkmark $       \\
        \textit{\textbf{VTAO (Ours)}} 
         & \textit{V+T+A} &  $\checkmark $   &  $\checkmark  $     &   $ \checkmark  $  \\
      \bottomrule
    \end{tabular}%
    \label{table:exp-Baselines}%
    \vspace{-0.4cm}
\end{table}%

\subsection{Effectiveness of pre-training with different modalities}
\begin{figure}[ht]
    \centering
    \vspace{+0.1cm}
    \begin{minipage}{0.5\textwidth}
        \centering
        \includegraphics[width=\linewidth]{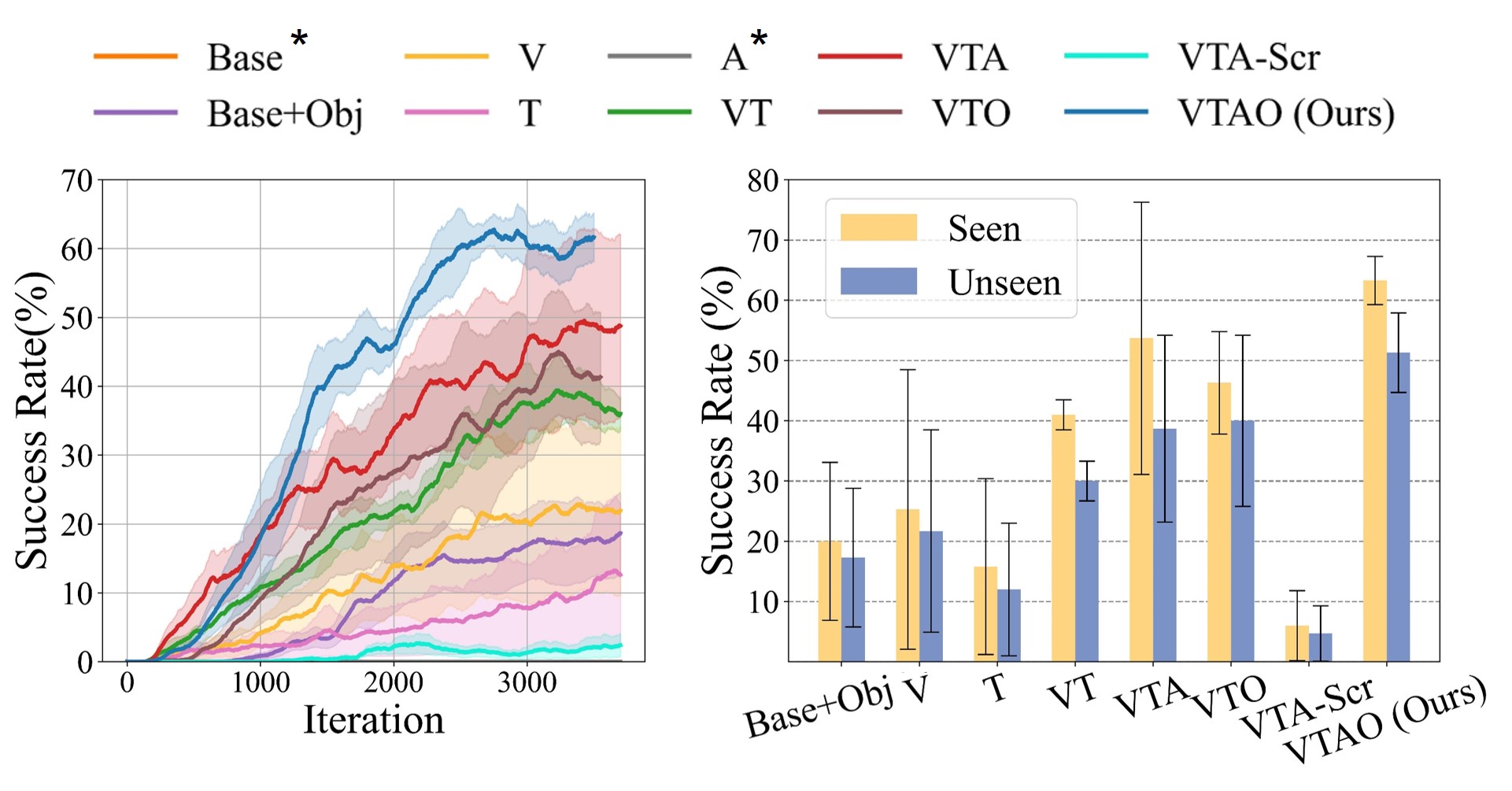}
        \captionsetup{font=small}
        \captionsetup{justification=raggedright, singlelinecheck=false}
        \vskip -0.3cm
        \caption*{* indicates that the method failed to train.}
    \end{minipage}
    \caption{Qualitative results of different pre-train methods in simulation. Left: training process; Right: evaluation results.}
    \label{fig:exp-MainExp}
    \vskip -0.4cm
\end{figure}

We design experiments to validate the effectiveness of our proposed multimodal joint pre-training. The experimental results are illustrated in the \figref[exp]{MainExp}. 

Compared to single-modality pretraining methods \textbf{V}, \textbf{T}, and \textbf{A}, as well as the multimodal method without pre-train \textbf{VTA-Scr}, the multimodal joint pretraining approach \textbf{VTA} demonstrates superior performance in downstream task manipulation success rates. This highlights the effectiveness of our multimodal joint pretraining strategy.

\szn{Analysis of modality integration reveals two key findings: First, the performance improvements from \textbf{VT} to \textbf{VTA} (incorporating action modality prediction) and to \textbf{VTO} (incorporating object understanding) demonstrate that both action prediction and object understanding significantly enhance visuo-tactile fusion.
Furthermore, \textbf{VTAO} achieves the highest success rates by combining both modalities, indicating complementary benefits between action prediction and object understanding tasks.}

\szn{Experiments with proprioception-only configurations (\textbf{Base} and \textbf{A}) fail to train, whereas \textbf{Base+Obj} achieves successful training. This underscores the critical importance of object state perception for enabling policy networks to dynamically adjust manipulation strategies based on target object status.} 

Although our method achieves the highest success rate, the overall performance (around 60\%) remains relatively low, which we attribute to the task's high complexity — specifically, the large variations in object shapes and sizes in our experimental setting and the high-dimensional action space of dual-hand manipulation.

\subsection{Ablation Study}

We conduct comprehensive ablation studies to validate key design choices in our approach. All experimental results are summarized in the Fig. 5. 

\subsubsection{\textbf{Effectiveness of Bimanual Pre-training}}
In our approach, we utilize tactile and action data from both hands for pre-training and downstream tasks. 
In this part, we validate the effectiveness of using both hands. 
We remove the left-hand action data in the \textbf{VTA} experiment (\textbf{v1}), resulting in version \textbf{v2}. Similarly, we remove the left-hand tactile data in the \textbf{VT} experiment (\textbf{v3}), producing version \textbf{v4}.
As shown in \tableref[exp]{Ablation}, using bimanual data consistently achieves higher success rates compared to using right-hand-only data, irrespective of the modality removed.
This indicates that although the left hand mainly performs basic grasping and holding, its inclusion significantly enhances coordination and improves success rates.


\begin{figure}[t!]
    \vspace{-0.2cm}
    \centering
    \begin{minipage}{\linewidth}
    \begin{table}[H]
    \tabcolsep=4pt
          \caption{Test success rates (\%) on seen and unseen bottles across different experimental settings.}
            \begin{tabular}{c|ccccc|cc}
                \toprule
                \textbf{Methods} &\textbf{Tac} & \textbf{Act} & \textbf{PredictAct}    & \textbf{ActToken} & \textbf{Obj} & \textbf{Seen}  & \textbf{Unseen} \\
                \midrule
                  \textbf{Ours}   & dual      &  dual       & 5steps &  $24 \times 2 $ & $ \checkmark $ & \textbf{63±4} & \textbf{51±7} \\
                \midrule
                  \textbf{v1}   & dual      &  dual       & 5steps & $ 24\times 2 $ & $ \times $     & \textbf{54±23} & \textbf{39±15} \\
                  \textit{v2} & dual      &  right      & 5steps & 24         & $ \times $  & 46±15 & 32±9 \\
                  \textit{v3} & dual      &  $ \times $    & $\times $ & $ \times $    & $ \times $ & 41±3  & 30±3 \\
                  \textit{v4} & right     &  $ \times $    &  $ \times $ & $ \times $  & $ \times $ & 37±17 & 20±7 \\
                \midrule
                  \textit{v5} & dual      &  dual       & NoPredict & $ 24\times 2 $ & $ \times $  & 35±25 & 21±16 \\
                  \textit{v6} & dual      &  dual       & 1steps      & $ 24\times 2  $ & $ \times $  & 44±8  & 17± 5 \\
                \midrule
                  \textit{v7} & dual      &  dual       & 5steps & $ 1\times 2 $  & $ \times $ & 31±4  & 18± 9 \\
                  \textit{v8} & dual      &  dual       & 5steps & $ 5\times 2 $  & $ \times $ & 50±16 & 26± 8 \\
                \bottomrule
            \end{tabular}%
          \label{table:exp-Ablation}%
    \end{table}
\end{minipage}

    \vspace{0.2cm}
    \hfill    

    \begin{minipage}{0.49\textwidth}
        \centering
        \includegraphics[width=\linewidth]{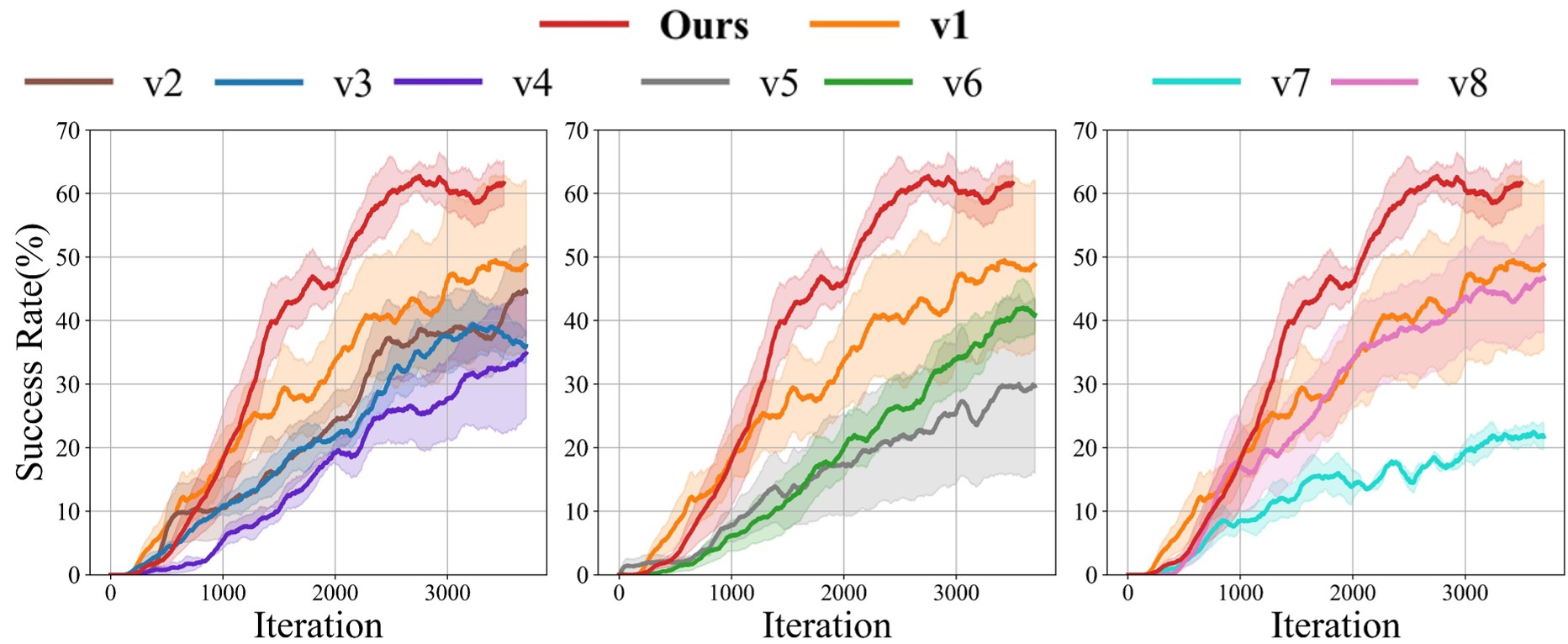}
        \caption{Training process for the ablation study. Left: Bimanual manipulation validation; Middle: Action prediction horizon ablation; Right: Ablation of the number of action tokens.}
        \label{fig:exp-Ablation}
    \end{minipage}
    \vskip -0.4cm
\end{figure}

\subsubsection{\textbf{Action Prediction Horizon}}
In our approach, we incorporate action prediction into the pre-trained model to forecast actions for the next 5 frames. In this part, we investigate the impact of the prediction horizon on downstream tasks. We evaluate the manipulation success rate of methods that either only reconstruct the action of the current frame (\textbf{v5}) or predict the action for the next frame (\textbf{v6}). The experimental results, as shown in \tableref[exp]{Ablation}, indicate that compared to the baseline without action addition (\textbf{v1}), solely reconstructing the action of the current frame has a detrimental effect. However, when action prediction was included, the success rate improved. 
This highlights the importance of predicting future actions in pretraining for robotic policies.

\subsubsection{\textbf{Number of Action Tokens~$\mathcal{N}$}}
Our method treats the action (angle values) of each joint in both hands as individual tokens, which are then fed into the Transformer. In this part, we explore different ways of segmenting the actions, considering hand actions based on each hand (\textbf{v7}) and each finger (\textbf{v8}). Experimental results, as shown in Table 2, demonstrate that setting the number of action tokens to 24×2 is more advantageous for VTA fusion.

\subsection{Real World Experiment}

\begin{figure}[h]
    \centering
    \includegraphics[width=0.9\linewidth]{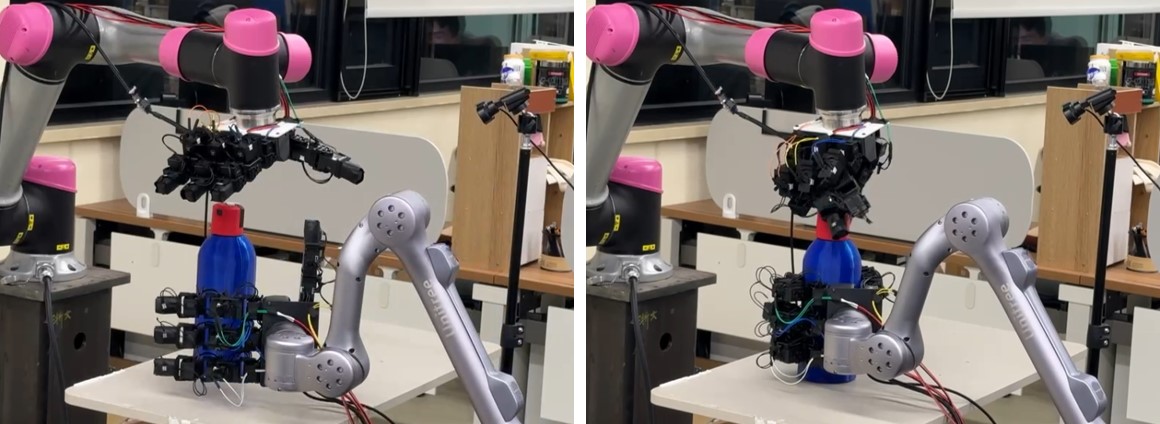}
    \caption{Our manipulation platform.}
    \label{fig:exp-RealWorldPlatform}
    \vspace{-0.5cm}
\end{figure}

We create a real-world experimental platform with dual Leap Hands~\cite{shaw2023leaphand} equipped with tactile sensors and a Unitree Z1 arm~\footnote{https://www.unitree.com/z1}, as shown in \figref[exp]{RealWorldPlatform}.
Visual perception is obtained through a ego-centric RGB camera, while tactile signals is captured via tactile sensors mounted on the dual Leap Hands and processed through binarization.
The simulation task for bottle cap unscrewing is recreated in Isaac Gym~\cite{makoviychuk2021isaac} using dual Leap Hands. 
The final validation confirms that our method supports effective physical experimentation.
Please refer to our submitted video for detailed demonstrations of the system's real-world performance.

\section{Conclusion}
This paper investigates the effectiveness of a pre-trained representation model integrating multi-modal input, action prediction, and object understanding for dexterous bimanual manipulation.
We design a novel pre-training framework, \method, which is pre-trained using human VTAO demonstrations.
We integrate a learning curriculum into RL to address the challenge of multi-skill learning within a bimanual dexterous skill.
The representation model is incorporated into the curriculum RL framework for bimanual manipulation.
Experimental results show that our approach significantly outperforms baseline methods. 
This study represents a pioneering effort in multimodal bimanual manipulation.
Future work may focus on: 
1) \szn{Scaling multimodal bimanual datasets across diverse scenes and tasks;} 
2) \szn{Extending to broader manipulation scenarios}; 
3) \szn{Exploring geometric object priors from vision models. Our method utilizes motion-captured object 3D bounding boxes (pose+size) for the object understanding. However, it may be replaced by poses estimated by large models for object pose estimaiton in the future.}


\bibliographystyle{IEEEtran}
\bibliography{mybibfile}






}
\end{document}